\documentclass{article}

\usepackage{arxiv}

\usepackage[utf8]{inputenc}	
\usepackage[T1]{fontenc}		
\usepackage{hyperref}			
\usepackage{url} 					
\usepackage{booktabs}			
\usepackage{amsfonts} 			
\usepackage{nicefrac}       		
\usepackage{microtype}      	
\usepackage{lipsum}

\usepackage{graphicx}
\usepackage{lineno}
\usepackage{natbib}
\usepackage{hyperref}
\usepackage{epsfig}
\usepackage{subfigure}
\usepackage{calc}
\usepackage{amssymb}
\usepackage{amstext}
\usepackage{amsmath}
\usepackage{mathtools}
\usepackage{multicol}
\usepackage{etoolbox}
\usepackage{enumitem}
\usepackage{diagbox}
\usepackage{float}
\usepackage{array}
\usepackage{adjustbox}

\modulolinenumbers[5]
\graphicspath{{./Figures/}}

\newcommand{\fold}[1]{\mathcal{#1}}
\newcommand{\set}[1]{\mathbb{#1}}
\newcommand{\vecx}[1]{\boldsymbol{#1}}

\hypersetup{
  pdfauthor = {},
  pdftitle = {},
  pdfsubject = {},
  pdfkeywords = {},
  colorlinks=true,
  linkcolor= blue,
  citecolor= blue,
  pageanchor=true,
  urlcolor = blue,
  plainpages = false,
  linktocpage
}

\newcommand{\secref}[1]{\autoref{sec:#1}}
\newcommand{\figref}[1]{\autoref{fig:#1}}
\newcommand{\eqnref}[1]{\autoref{eqn:#1}}
\newcommand{\tabref}[1]{\autoref{tab:#1}}

\newcommand{\eg}{\emph{e.g.}}
\newcommand{\ie}{\emph{i.e.}}

\title{Ensuring Learning Guarantees on Concept Drift Detection with Statistical Learning Theory}

\author{Lucas~de~C.~Pagliosa, Rodrigo~F.~Mello
\thanks{L. Pagliosa and R. Mello were with the Institute of Mathematics and Computer Science, University of S\~ao Paulo, 400, 13566-590, S\~ao Carlos, Brazil e-mail:lucas.pagliosa@usp.br, mello@icmc.usp.br.}}

\begin{document}
\maketitle

\begin{abstract}
Concept Drift (CD) detection intends to continuously identify changes in data stream behaviors, supporting researchers in the study and modeling of real-world phenomena. Motivated by the lack of learning guarantees in current CD algorithms, we decided to take advantage of the Statistical Learning Theory (SLT) to formalize the necessary requirements to ensure probabilistic learning bounds, so drifts would refer to actual changes in data rather than by chance. As discussed along this paper, a set of mathematical assumptions must be held in order to rely on SLT bounds, which are especially controversial in CD scenarios. Based on this issue, we propose a methodology to address those assumptions in CD scenarios and therefore ensure learning guarantees. Complementary, we assessed a set of relevant and known CD algorithms from the literature in light of our methodology. As main contribution, we expect this work to support researchers while designing and evaluating CD algorithms on different domains.
\end{abstract}

\keywords{Concept Drift Detection \and Statistical Learning Theory \and Probabilistic Learning Guarantees \and Drift Detection Guarantees \and Supervised Learning}


\section{Introduction}
\label{sec:intro}

Data streams are seen as open-ended sequences of uni or multidimensional observations rather than batch-driven datasets~\citep{lichman:uci:13}\footnote{In the context of this paper, we approach unidimensional streams, however our strategy can be extended to address multiple dimensions, without loss of generality as performed in~\citep{serra:09:njp}.}. Those observations are generated by processes modeled as stochastic and/or chaotic dynamical systems~\citep{kantz:97:book}, which simulate several phenomena at different periods of time~\citep{agarwal1995dynamical,Metzger1997,RIOS201511}. Nevertheless, those processes or their parameters may eventually change over time, such as how a disease/medicine impacts on someone's blood temperature~\citep{Andrievskii2003}. Those cases, hereinafter referred to as Concept Drifts (CD), are important to be considered while modeling especially chaotic data, whose system present recurrent behaviors. Later, specialists can analyze those anomalous and decisive instants for further comprehension on studied phenomenon.

In practice, CD algorithms compare features from current to next observations in order to detect relevant changes~\citep{gama:14:acm,lu:kde:18}. Such features are usually represented by classification performances~\citep{gama:sac:04,EDDM,bifet:sigkdd:09} or by statistical measurements~\citep{gama:14:acm,page:bio:54,bifet:sigkdd:09}. Moreover, despite former methods generally lead to more robust results, they demand class labels to perform supervised learning, which may be impractical when dealing with data continuously collected over time. Conversely, statistical-based methods have the advantage of requiring no label, but their simplistic models usually is not enough to properly distinguish general processes, especially when dealing with non-stationary and chaotic phenomena~\citep{costa:eswa:16}. 

Besides relevant contributions, both branches do not provide learning guarantees to support CD detection, although some authors claim that performance measures such as Mean Time between False Alarms (MTBFA), Mean Time for Detection (MTD) and Missed Detection Rate (MDR)~\citep{costa:eswa:17} can ensure such commitment, \eg, in terms of accuracy. In that sense, we remind the reader that those measures cannot be trustworthy when the algorithm poorly (under or overfitting) generalizes data. In an extreme case, a CD algorithm that randomly issues drifts (such as flipping a coin) or whose model memorized all training data may still provide adequate performances according to those measurements. Thus, instead of considering specific measurements on particular scenarios, we propose a general and formal approach to ensure CD detection relying on the Statistical Learning Theory (SLT), proposed by~\citet{vapnik:98:book}. In summary, our strategy provides the necessary probabilistic foundation to ensure learning while analyzing data streams. As consequence, drifts are not reported by chance, and validation methods can be fully employed later on. 

According to the SLT, learning occurs when the empirical risk (computed over the training data) does not significantly differ from the expected risk (inferred over the whole population) \emph{and} the empirical risk is small enough according to the specialist~\citep{mello:book:18}. In order to formulate such a theoretical framework, Vapnik had to employ a strategy to prove the convergence of the adopted classifier to the best as possible model inside the algorithm bias, which has motivated him to consider the Law of Large Numbers (LLN)~\citep{devroye:96:book} in this process. However, in order to employ the LLN, Vapnik had to follow a set of assumptions, in which two of them are especially controversial in CD scenarios: (i) input data must be sampled in an independently and identically distributed (i.i.d.) manner; and (ii) the Joint Probability Distribution (JPD), mapping the relationship between input and output spaces, must be static/fixed.

This controversy firstly arises from the fact that real-world data streams typically present time-dependent observations, given they represent the same phenomenon over different timestamps. Secondly, if a CD is supposed to happen, some changes over the JPD are also expected to occur. Even so, given that the SLT is the most complete framework to ensure learning for supervised algorithms (we are not aware of any other theoretical basis to tackle the problem in the same level), we tried to address such drawbacks by proposing a set of adaptations to satisfy both requirements. More precisely, our goal in this manuscript is to elaborate the necessary conditions a CD algorithm should satisfy to ensure learning while reporting drifts. It is worth to make clear, however, that we do not intend to propose a new algorithm in the process. Nonetheless, we have analyzed some of the CD literature in light of our theoretical point of view to show that many of them are, in fact, in disagreement with the SLT. Thus, this does not mean those algorithms will not work and generate fair results, but that no theoretical guarantee could be derived from their application. Lastly, after employing our strategy to ensure learning bounds, other performance measures can be safely used to validate the quality of reported drifts.

The remaining of this manuscript is structured as follows. \secref{concepts-and-nomenclature} introduces the background involving CD algorithms, SLT and the main concepts from the area of Dynamical Systems. \secref{related-work} shows the related work on theoretical guarantees in CD problems. \secref{learning-in-cd-scenarios} discusses our proposed methodology to ensure learning bounds while tackling the concept drift scenario. \secref{cd-algorithms-according-to-our-methodology} describes important algorithms on concept drift detection, highlighting when they satisfy (or not) Vapnik's assumptions. Concluding remarks are drawn in \secref{conclusions}.

\section{Concepts and Nomenclature}
\label{sec:concepts-and-nomenclature}

This section introduces the CD nomenclature used in this manuscript. In addition, Dynamical System concepts, such as phase spaces reconstruction, are briefly covered. Lastly, a short description of SLT and its assumptions are given. Despite not commonly found in the concept-drift literature, the two latter topics must be presented in order to explain our methodology.

\subsection{Concept Drift Detection}
\label{sec:concept-drift-concepts}

\begin{figure}[htb]
\centering
\includegraphics[width=0.65\linewidth]{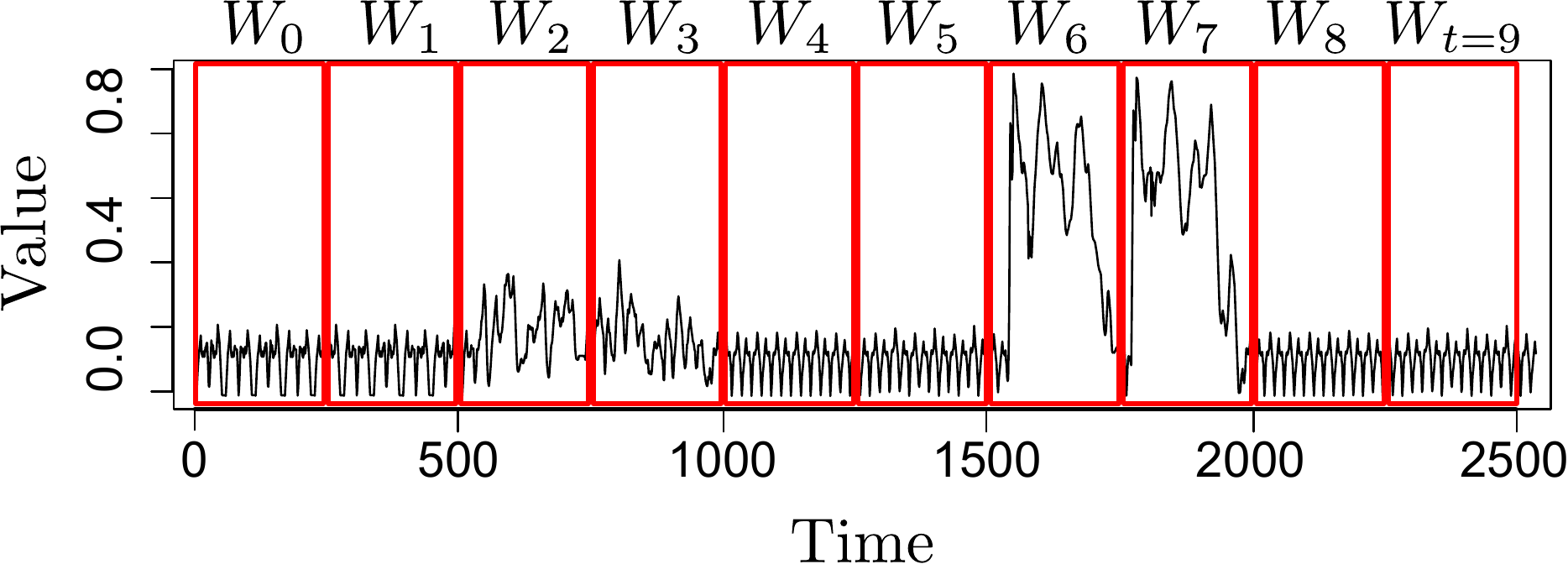}
\caption{A data stream divided in $10$ windows (red boxes) with no overlapping, each containing $n = 250$ observations.}
\label{fig:data-stream-windows}
\end{figure}

Let a data stream $\fold{D}$ be defined as the sequence of observations:
\begin{equation}
\fold{D} = \{x(0), x(1), x(2), \cdots, x(\infty)\}, ~~ x(k) \in \set{R},
\label{eqn:data-stream}
\end{equation}
describing the behavior of some phenomenon along time. Moreover, a data stream defines a continuous flow of incoming data, whose observations are derived from (potentially) multiple Joint Probability Distributions (JPDs). Thus, given a data window $W_i$ representing the $i$th set of observations in a specific interval of time, we have that ${\bigcup_{i=0}^{t \rightarrow \infty} W_i = \fold{D}}$ (see \figref{data-stream-windows}), where $t$ is the current timestamp.
Additionally, despite the configuration of windows may vary from application to application, it is common to assume a fixed length $n$ for every window without the overlapping of observations, such that:
\begin{equation}
W_i = \{x(i \times n), \cdots, x((i \times n) + n - 1)\}.
\label{eqn:window}
\end{equation} 

Thus, if $s$ represents the timestamp from which observations started to be collected from some phenomenon (initially set to zero), a CD algorithm induces the indicator function:
\begin{equation}
g_t: \phi(f_t) \rightarrow [0, 1],
\label{eqn:phenomenon-model}
\end{equation}
which basically classifies whether the current window $W_t$ continues to represent the same phenomenon of past windows or not. To avoid redundancy, the index of $g$ is omitted unless explicitly necessary to express a specific time (\secref{related-work}). Moreover, function $\phi$ extracts the vector of features $\vecx{v}_i = \phi(f_i)$ from the model:
\begin{equation}
f_i: \fold{X}_i \rightarrow \fold{Y}_i,\; \forall i \in [s, t],
\label{eqn:window-model}
\end{equation}
where $\fold{X}_i$ and $\fold{Y}_i$ are the input and output (class) spaces of window $W_i$, respectively, derived either after applying dynamical system reconstructions or by using statistical measurements (\secref{adapting-slt-to-cd-scenarios}). Those spaces are necessary in any CD algorithm to perform supervised learning, and more details about both are given in \secref{adapting-slt-to-cd-scenarios}. Complementary, $\vecx{v}_i$ can simply be the result of $f_i$ itself so that $\phi$ is the identity function (\eg, if the model is based on the mean, variance, or entropy of the window) or, under more sophisticated scenarios, $\phi$ can be a more complex kernel function (\eg, if $f_i$ is inferred from an Artificial Neural Network~\citep{haykin:09:book}, features could be given by unit weights or the results of activation functions).

Formally, we define $\vecx{v}_t$ as the features obtained for the current window $\fold{X}_t$, and $\vecx{v}_{[s,t)}$ as the set of features extracted from past windows $\fold{X}_{[s,t)}$. In this context, the CD algorithm, here responsible for inducing function $g$, reports a drift whenever $\vecx{v}_{t}$ significantly differs from $\vecx{v}_{[s, t)}$ by more than an acceptable threshold $\lambda$. On the other hand, if the divergence between features is acceptable, then $g$ infers that $\vecx{v}_t$ and $\vecx{v}_{[s, t)}$ belong to the same phenomenon and the analysis should be carried on to next windows. At this step, $g$ has two options: either it updates the set of features in the form ${\vecx{v}_{[s, t]} = \vecx{v}_{[s, t)} \cup \vecx{v}_t}$ or it merely sets ${\vecx{v}_{[s, t)} = \vecx{v}_t}$ ($t$ is naturally incremented after both cases). Note, however, that in the former scenario, $\vecx{v}_{[s,t)}$ is much greater than $\vecx{v}_t$. Thus, $g$ must either perform aggregations or apply kernel functions to make sure that $\vecx{v}_{[s,t)}$ has the same number of features than $\vecx{v}_t$, so a fair comparison between them is made. In this context, the following strategy is commonly used to compare them:
\begin{align}
g(\vecx{v}_t) = \left\{ \begin{array}{cl}
1,	& \text{if} ~~ \vecx{v}_t - \left( \mu_{\vecx{v}_{[s, t)}} + \eta\sigma_{\vecx{v}_{[s, t)}} \right) > \lambda\\
	& \text{or} ~~ \vecx{v}_t - \left( \mu_{\vecx{v}_{[s, t)}} - \eta\sigma_{\vecx{v}_{[s, t)}} \right) < -\lambda,\\
0, 	& \text{in case of no drift},
\end{array} \right.
\label{eqn:alert-drift}
\end{align}
where $\mu_{\vecx{v}_{[s, t)}}$ and $\sigma_{\vecx{v}_{[s, t)}}$ are the average and standard deviation of past features, and $\eta \in \set{R}_+$ controls the detection sensitiveness.

\subsection{Dynamical Systems}
\label{sec:dynamical-systems}

This section describes Dynamical System concepts necessary to automatically derive input $\fold{X}_i$ and output $\fold{Y}_i$ from window $W_i$. Despite those sets can be generated in different ways, by means of Fourier coefficients~\citep{bracewell:78:book} or statistical measurements, for instance, we suggest them to be derived from states in the phase space for a robust modeling, as we explain next. Furthermore, this section presents some concepts not commonly adapted for the CD literature. Still, they are important to satisfy SLT assumptions according to our point of view.

A dynamical system $\fold{S}^d = \{\rho_0, \rho_1, \cdots\}$ is composed of a set of $\fold{D}$-dimensional states that, driven by a generating (a.k.a. governing) rule $R(\cdot)$, models the behavior of some phenomenon in function of state trajectories, such that:
\begin{equation}
R: \fold{S}^d \rightarrow \fold{S}^d,
\end{equation}
where $d$ corresponds to the number of degrees of freedom the system has, \ie, the number variables/dimensions needed to describe $R(\cdot)$. 

\begin{figure}[htb]
\centering
\includegraphics[width=0.65\linewidth]{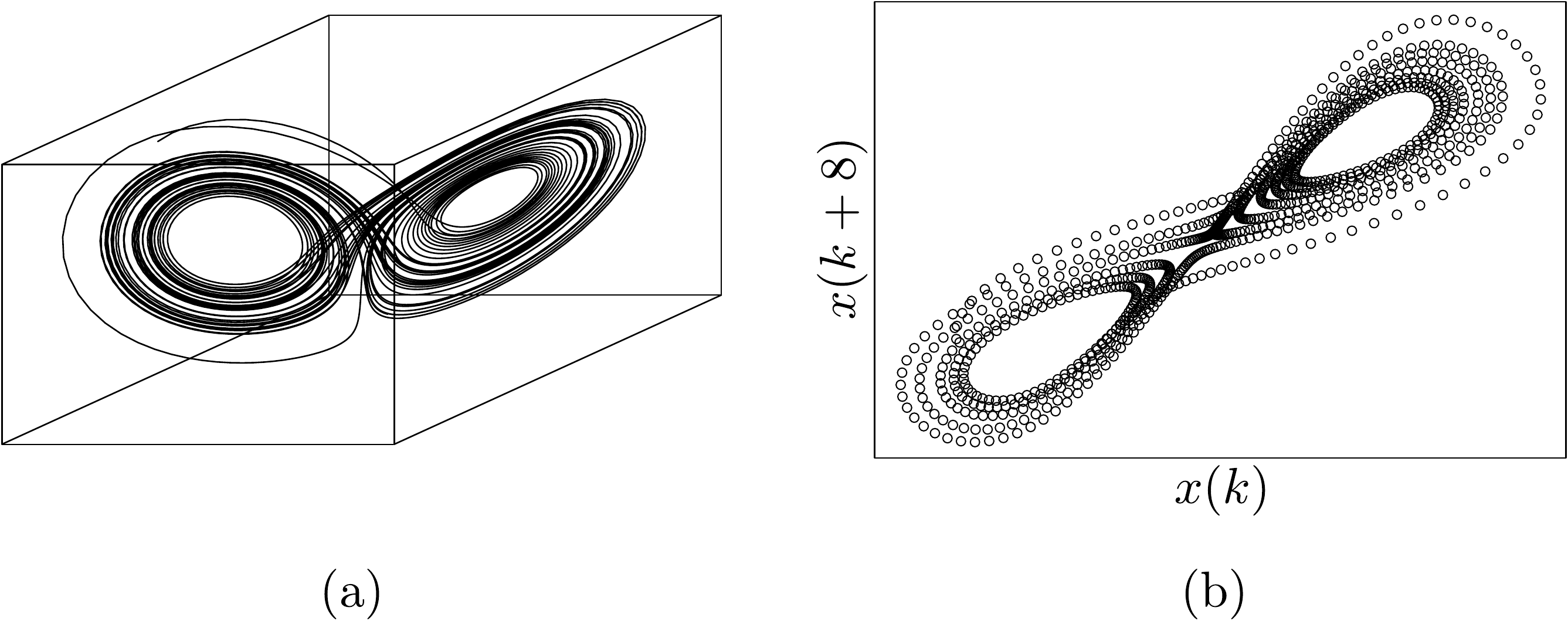}
\caption{(a) Continuous Lorenz attractor. (b) Discrete reconstruction of the Lorenz attractor using $(m, \tau) = (2, 8)$. As it can be noticed, no matter the order of how phase states are drawn, the attractor is still depicted the same way, which reinforces the fact that states are independent from each other.
}
\label{fig:lorenz-attractor}
\end{figure}

Moreover, when the number of states are enough to represent all system dynamics, than $\fold{S}^d$ is referred to as the \emph{phase space} $\fold{P}$ in which the variable \emph{time} is no longer explicitly required for modeling, given all possible trajectories are bound by the structure of a potentially low-dimensional manifold known as \emph{attractor}~\citep{alligood:96:book}. \figref{lorenz-attractor}(a) illustrates such structure for the well-known $3$-dimensional Lorenz system~\citep{tucker:1999}.
As consequence, phase states become independent so they can be identically distributed, resulting in satisfying the i.i.d. assumption necessary to perform the SLT~\citep{pagliosa:eswa:17}.

When dealing with data streams symbolizing observations collected from a single dimension of such phase space, we rely on the embedding theorem proposed by~\citet{takens:1981}, a direct extension from Whitney's studies in the area of differential manifolds~\citep{whitney:1936}, to reconstruct $\fold{S}^d$ from $\fold{D}$. This embedding theorem has been widely employed and confirmed to be the most adequate to reconstruct different types of dynamical systems~\citep{ravindra:1998}. As proposed, a time series (finite sequence of $n$ observations) in the form ${T = \{x_0, x_1, \cdots, x_{n - 1}\}}$ can be reconstructed into a set of $N$ states, whose structure $\fold{P}$ is a manifold diffeomorphic ($\approx$) to $S^d$, if states $p_{k \in [0, N - 1]}$ are in the form:
\begin{equation}
\rho_k(m,\tau) = (x_k, x_{k+\tau}, \cdots, x_{k+(m-1)\tau}) \in \fold{P} \approx \fold{S}^d,
\label{eqn:embedding}
\end{equation}
in which $m$ corresponds to the embedding dimension and $\tau$ is the time delay. \figref{lorenz-attractor}(b) shows the result of the reconstruction for the Lorenz data~\citep{tucker:1999}, using $m = 2$ and $\tau = 8$. In our context, $T$ corresponds to some window $\fold{W}_{i \in [0,t]}$ (\eqnref{window}) such that ${x_k = x((i \times n) + k)}$.

\subsection{Statistical Learning Theory}
\label{sec:slt}

In the context of supervised learning, data is divided as input and class spaces, represented as $\fold{X} = \{x_1, \cdots, x_n\}$ and $\fold{Y} = \{y_1, \cdots, y_n\}$, respectively, whose instances are sampled from a Joint Probability Distribution (JPD) ${P(\fold{X} \times \fold{Y})}$ and, then, employed to induce some learning model ${f: \fold{X} \rightarrow \fold{Y}}$ which is expected to provide the smallest as possible error or loss. Based on such definition, the Statistical Learning Theory (SLT)~\citep{vapnik:98:book} provides a theoretical framework to ensure the model $f$ will generalize similarly on unknown examples as it performs on known data. As consequence, SLT supports to find good estimates for the expected value of some loss function, allowing to infer $f$ based on performance measurements computed on known (training/test) samples.

Given a problem and its representative loss function $\ell$,~\citet{vapnik:98:book} defined the empirical risk, \ie, the error computed over some sample, as:
\begin{equation}
R_{\text{emp}}(f) = \frac{1}{n} \sum_{k=1}^{n} \ell(x_k, y_k, f(x_k)),
\label{eqn:empirical-risk}
\end{equation}
and the expected (a.k.a. real or actual) risk, which is computed by assessing the whole JPD $P(\fold{X} \times \fold{Y})$), in the form:
\begin{equation}
R(f) = \text{E}(l(\fold{X}, \fold{Y}, f(\fold{X}))),
\label{eqn:true-risk}
\end{equation}
being $\text{E}$ the expected value. From that context, Vapnik relied on the Law of Large Numbers (LLN)~\citep{revesz:67:book} to prove the following:
\begin{equation}
P(|R(f) - R_{\text{emp}}(f)| \geq \epsilon) \rightarrow 0, \;\; n \rightarrow \infty,
\label{eqn:ermp}
\end{equation}
\ie, the empirical risk of a classifier $R_{\text{emp}}(f)$ probabilistically converges to the expected risk $R(f)$ as the sample size $n$ tends to infinity, considering $R(f)$, $R_{\text{emp}}(f)$, $\epsilon \in [0,1]$ and some assumptions are satisfied (as described next). From such statement, a.k.a. Empirical Risk Minimization Principle (ERMP), Vapnik defined that a model $f$ generalizes when the difference ${|R(f) - R_{\text{emp}}(f)|}$ approaches zero, and learns when $f$ generalizes \emph{and} its empirical risk $R_{\text{emp}}(f)$ is small enough according to the specialist.

As \eqnref{true-risk} cannot be calculated in practice, since it demands an infinite number of observations to analyze the whole joint space, Vapnik used the Symmetrization Lemma~\citep{devroye:96:book} to quantify learning in real scenarios as follows:
\begin{align}
P \left( \sup_{f \in \fold{F}} | R(f) - R_\text{emp}(f) |\geq \epsilon \right) &\leq \nonumber\\
2P \left( \sup_{f \in \fold{F}} |R'_{\text{emp}}(f) - R_{\text{emp}}(f) |\geq \frac{\epsilon}{2} \right) &\leq \delta, \;\; n \rightarrow \infty,
\label{eqn:symmetrization-lemma}
\end{align}
where $\delta$ is the acceptable confidence level for the probabilistic measure, $\fold{F}$ is the set containing all functions an algorithm is capable of admit, a.k.a. the algorithm bias, and $R'_{\text{emp}}(f), R_{\text{emp}}(f) \in [0,1]$ are the empirical risks of two different samples having the same size $n$. In summary, this lemma states that if two independent samples have empirical risks that do not diverge more than $\epsilon$ as $n$ increases, than it is expected the empirical risk to be a good estimator for the expected risk.

Finally, it is worth to mention that $\fold{F}$ must be in parsimony with the Bias-Variance Dilemma (BVD)~\citep{luxburg:11:book}, otherwise \eqnref{symmetrization-lemma} is inconsistent. In other words, if the class of functions in $\fold{F}$ has weaker bias (less restrictive), then $\fold{F}$ contains many more distinct functions to represent some training set, what generally leads to overfitting. Conversely, if $\fold{F}$ has a strong bias (more restrictive), fewer functions compose such a space, making it prone to underfitting. Thus, a balanced complexity of the function class is recommended to achieve the best as possible risk minimization~\citep{geman:92:nc}. This is important because two underfitted models, for instance, might still provide similar empirical errors as they both learned from the average, but they will not generalize on future data.

While elaborating the SLT, Vapnik took advantage of the LLN to formulate and prove learning bounds for ERMP and the Symmetrization Lemma. Due to his formulation, the learning guarantees are only held if the following assumptions are satisfied: examples must be independent from each other and sampled in an identical manner (A1); no assumption is made about the joint probability distribution (JPD), otherwise one could simply estimate its parameters (A2); labels can assume non-deterministic values due to noise and class overlapping (A3); the JPD is fixed, \ie, it cannot change over time (A4); and, finally, data distribution is still unknown at the time of training, thus it must be estimated using data examples (A5).

It is simple to observe that assumptions A2, A3 and A5 are straightforward, since they define most of real-world scenarios. However, assumptions A1 and A4 are more difficult to hold especially in (but not limited to) the CD scenario, in which observations are time dependent and different phenomena (with distinct JPDs) are expected to happen as widely discussed in the CD related work~\citep{gama:14:acm}. Nevertheless, even with such controversies, we still choose to rely on the SLT due to its robust framework towards probabilistic convergence in supervised learning. Moreover, we can adapt the CD algorithm to satisfy SLT assumptions, as we show in \secref{learning-in-cd-scenarios}.

\section{Related Work}
\label{sec:related-work}

Regarding the task of learning in drifting concepts, we reinforce that validation methods (\eg, accuracy, false positive/negative rates) are insufficient to draw conclusions on that aspect. Despite a good way to measure the quality of reported drifts on a limited number of observations, those metrics do not give any probabilistic guarantee that the reported performances will continue over time. Regardless, there are a few studies in the literature aiming to support theoretical learning~\citep{tsymbal:04}. For instance~\citet{kuh:nips:1991} relied on PAC-Learning~\citep{valiant:acm:1984} to estimate the minimum window length necessary to trust in reported drifts, whereas~\citet{helmbold:ml:1994} proposed a weak upper bound to delimit how fast drifts should occur based on the same framework. As consequence, they both conclude learning could be only ensured if concept changes occurred ``slowly'' enough according to the window length.

With respect to those articles, we highlight that despite PAC-Learning contributions, such as the introduction of computational complexity theory in Machine Learning, the SLT consists of a much more complete and robust framework towards supervised learning. Even so, there are equivalences between both, such that if the algorithm bias $\fold{F}$ is PAC-learnable, than the Vapnik-Chervonenkis (VC)-dimension~\citep{luxburg:11:book, mello:book:18} of $\fold{F}$ should be finite\footnote{Further comments on the VC-dimension are out of the scope, see cited references for details.}. In fact, the upper and lower bounds formulated in the previously-cited articles explicitly consider the VC-dimension in their equations. Therefore, as the VC-dimension is a direct consequence of the ERMP (\eqnref{ermp}), one should firstly be in accordance with the SLT requirements (by using our methodology, for instance) in order to rely on such bounds. Lastly, we admit that the VC-dimension is a difficult measurement to compute in practice, what might invalidate both approaches for general cases.

More recently,~\citet{mello:yuli:18} elaborate a framework to ensure unsupervised learning guarantees based on Algorithmic Stability (AS)~\citep{bousquet:jlmr:02}, which presents conditions for the probabilistic convergence between an arbitrary function and its expected value using the McDiarmid’s Inequality~\citep{mcdiarmid:sic:1989}. However, as the authors elaborated themselves, ``the SLT has a strong connection to the whole theory employed to ensure Algorithmic Stability''~\citep{mello:yuli:18}. As it can be noticed, the comparison between the current and past windows, proposed by AS, is related to the Symmetrization Lemma (\eqnref{symmetrization-lemma}), and most of the used inequalities are based on the Law of Large Numbers (LLN), which also requires input data to be i.i.d. Therefore, our methodology could also be applied on their framework.

As a conclusion, we observe that the literature lacks of supervised learning guarantees in CD scenarios, what have motivated us to associate the SLT towards CD detection. In that sense, our methodology comes as a novel but simple approach to ensure reported drifts are not by change, but due to actual changes on data behaviors.

\section{Ensuring Learning in Concept Drift Scenarios}
\label{sec:learning-in-cd-scenarios}

In this section, we translate some of the presented SLT bounds (\secref{slt}) towards the CD context. From that, we elaborate the necessary conditions a CD algorithm should satisfy to meet such theoretical framework.

\subsection{Adapting the SLT to CD scenarios}
\label{sec:adapting-slt-to-cd-scenarios}

In this section, we make a summary of all previously presented concepts to introduce our methodology. Firstly, we remind the reader that CD algorithms typically assume the learning model ${f_i: \fold{X}_i \rightarrow \fold{Y}_i}$ is inferred over some window $W_i$, where $\fold{X}_i$ corresponds to the input examples, and $\fold{Y}_i$ is associated with the respective classes. Those classes are usually not available as it is difficult for a specialist to continuously label observations collected over time, most especially for high-frequency streams.

From that, we conclude that class labels must be somehow devised from the data stream itself in an online fashion. As consequence, the input example might also change according to the chosen procedure to determine classes. From that, two possible strategies have been used to tackle such an issue: (i) if $f_i$ is result of a regression performed on the phase space, then each input $x_k \in \fold{X}_i$ is a tuple composed of the first $(m - 1)$ dimensions of $\rho_k$, while the respective class label $y_k \in \fold{Y}_i$ is the value of the $m$th dimension, as show in \tabref{tabular-logistic}; and (ii) when the class information is the simple result of a measurable function $m(\cdot)$ computed over the data window, such as the average, variance, kurtosis, etc., then $\fold{X}_i = W_i$ (the input is simply the window) and its output comprises the label $m(\fold{X}_i) = y_{[1,n]} \in \fold{Y}_i$~\citep{bifet:sigkdd:09}.
\begin{table}[htb]
\renewcommand\arraystretch{1.2}
\setlength{\tabcolsep}{6pt}
\begin{center}
\caption{Input and output spaces for the Lorenz system, embedded with $m=2$ and $\tau=8$ (\eqnref{embedding}) and illustrated in \figref{lorenz-attractor}(b). In the example, the phase space was reconstructed from the window $W_i$ but, for clarity, its states were represented as a generic time series, such that ${x_k = x((i \times n) + k)}$.}
\label{tab:tabular-logistic}
\begin{tabular}{|c|c|c|}
\hline
States & $\fold{X}_i$ & $\mathcal{Y}_i$ \\
\hline
$\rho_0 = (x_0, x_7)$  			    & $x_{0}$		& $x_{7}$   \\ \hline
$\rho_1 = (x_1, x_8)$ 				& $x_{1}$		& $x_{8}$   \\ \hline
\vdots                 				& \vdots 		& \vdots    \\ \hline
$\rho_{N-2} = (x_{N-2}, x_{N+5})$ 	& $x_{N-2}$	& $x_{t+5}$     \\ \hline
$\rho_{N_1} = (x_{N-1}, x_{N+6})$ 	& $x_{N-1}$	& $x_{t+6}$     \\ \hline
\end{tabular}
\end{center}
\end{table}

In the next step, function $\phi$ extracts the vector of features $\vecx{v}_i$ from the inferred model $f_i$, such that $\vecx{v}_i = \phi(f_i)$. For instance, if $f_i$ is in the form of a neural network, then the empirical risk or the trained weights might represent $\vecx{v}$. The indicator function $g$ is then responsible for mapping every feature vector into a binary space, indicating whether a drift has happened given the current data window (\eqnref{phenomenon-model}). If no drift is detected, then $g$ can either change its current model to the newest (${\vecx{v}_{[s, t)} = \vecx{v}_t}$) or update it based on the new features (${\vecx{v}_{[s, t]} = \vecx{v}_{[s, t)} \cup \vecx{v}_t}$). As we show, the latter option is the best as $g$ continuously approximates the true set of features corresponding to the analyzed phenomenon, allowing us to elaborate the following connection with the ERMP (\eqnref{ermp}):
\begin{equation}
P(\lVert \vecx{v}_{[t, +\infty]} - \vecx{v}_{[s, t)} \rVert \geq \epsilon) \rightarrow 0, \;\; t \rightarrow \infty.
\label{eqn:ermp-cd}
\end{equation}
In other words, if the difference between the empirical and true risk decreases as the sample size increases, thus the features extracted over time should also converge to the features computed over the entire data population. Ideally, we should set some window length large enough to contain all observations from the analyzed phenomenon. However, this becomes a great challenge since: (i) we do not have access to all observations from the phenomenon, and (ii) several drifts are expected to happen in early windows. Thus, we decided to adapt the Symmetrization Lemma (\eqnref{symmetrization-lemma}) to represent learning in terms of windows features, in the form:
\begin{align}
P \left( \sup_{f_i \in \fold{F}} \lVert \vecx{v}_{[t, +\infty]} - \vecx{v}_{[s, t)} \rVert \geq \epsilon \right) &\leq \nonumber\\
2P \left( \sup_{f_i \in \fold{F}} \lVert \vecx{v}_t - \vecx{v}_{[s, t)} \rVert \geq \frac{\epsilon}{2} \right) &\leq \delta, \;\; t \rightarrow \infty,
\label{eqn:symmetrization-lemma-cd}
\end{align}
remembering that $\vecx{v}_{[s, t)}$ represents an aggregation of all measurements for past windows, such that the sample size of $\vecx{v}_t$ and $\vecx{v}_{[s, t)}$ is the same. Therefore, if such a difference is held as new windows are processed, we have probabilistic support that $g$ is actually learning from data.

\subsection{Satisfying SLT assumptions}
\label{sec:satisfying-slt-assumptions}

In order to rely on \eqnref{symmetrization-lemma-cd}, however, we must satisfy assumptions A1 and A4 listed in \secref{slt}. Moreover, we also need to ensure such equation is consistent by choosing a CD algorithm whose complexity is in parsimony with the BVD~\citep{luxburg:11:book}.

Firstly, we draw attention to the fact that, as we propose, drifts will occur only between windows, not among observations. According to our approach, the algorithm responsible for inferring $f_i$ is expected to deal with A1, while the model $g$ faces the challenge in A4. Therefore, models $f_i$ should employ some strategy to map observations into a different space, ensuring data becomes i.i.d. For instance, the Fourier transform~\citep{bracewell:78:book} could map windows into the frequency space, or the Takens' embedding theorem~\citep{takens:1981} may be used to map such observations into phase spaces~\footnote{We encourage the reader to use this option, as the reconstruction of phase spaces allows a more robust data analysis.}. Complementary, model $g$ assumes that each data window may come from distinct but fixed/unique probability distributions, so when this indicator function reports a drift, any previous model should be discarded, allowing a fresh start to analyze a next coming distribution while still ensuring learning guarantees.

Regarding under/overfitting, one should choose functions $f_i$ and $g$ whose bias complexity is considered moderate according to the BVD~\citep{luxburg:11:book}. When $f_i$ is based on statistical measures, usually the search space consists of a single function, making $f_i$ more prone to underfitting. Further, such model is only effective to test particular hypotheses, \eg, when data is statistically stationary (which we claim it is unlikely to happen when dealing with real, nonlinear and/or chaotic datasets).

Alternatively, when $f_i$ is inferred based on Dynamical System approaches, the model usually relies on the distances among phase states based on an open-ball radius $\varepsilon$, given the topological space of attractors~\citep{tu:book:10}. In those cases, if $N$ is the number of phase states, models using a small $\varepsilon$ typically overfit, as $f_i$ memorizes each state. On the other hand, the use of an excessively large radius makes the model learns from the attractor/space average, leading to underfitting~\citep{mello:book:18}. Thus, a balanced-complexity model should be based on a fair and adaptive percentage of distances among states, \eg, $\varepsilon$ can be defined in terms of the open ball containing the $K$ nearest neighbors or some distance quantile computed on the phase space. Regarding the indicator function $g$, the comparison between windows should follow some strategy as the one defined in \eqnref{alert-drift}, otherwise simpler functions would lead to underfitting and more complex indicators to overfitting.

In summary, our methodology to satisfy SLT requirements in CD scenarios is composed as follows: (R1) the indicator function $g$ should be updated based on past data, so that the underlying phenomenon is better represented; (R2) the model $f_i$ must receive i.i.d. data, something ensured by a different pre-processing approach (\eg, Fourier transform or phase space reconstruction); (R3) the function $g$ should compare features from the same JPD. Otherwise a reset in $g$ is necessary; (R4) the algorithm bias from both $g$ and $f_i$ should be in parsimony with the BVD.

\section{Concept Drift Algorithms According to our Methodology}
\label{sec:cd-algorithms-according-to-our-methodology}

As discussed in \secref{concept-drift-concepts}, a CD algorithm can be divided in two components: (i) the first responsible for extracting features from data windows, using function $f_i$; and (ii) the second capable of comparing those features using some indicator function $g$. In light of such perspective, we present state-of-the-art algorithms and highlight how they approach requirements R1--R4. Firstly, the Cumulative Sum (CUSUM)~\citep{page:bio:54} algorithm reports a drift whenever an incoming observation is significantly different from the sum of past data. Thus, knowing that $g_s$ is initially set to zero, a drift occurs when:
\begin{equation}
g_t = \max(0, g_{t - 1} + x(t)) \geq \lambda,
\label{eqn:cusum}
\end{equation}
in which $\lambda \in \set{R}^+$ (here and next) is an acceptable threshold, and $x(t)$ consists in a single observation such that $\fold{X}_i = x(i)$ (window length $n = 1$). In this scenario, $f_i: x(i) \rightarrow x(i)$ and $\phi(f_i) = x(i)$ correspond to the identity function while $g_t$ is a model directly correlated to the average of such a phenomenon. A drift is reported when $g_t$ results in a value greater than the threshold $\lambda$, and $g_{s = t}$ resets the analysis for a new phenomenon (satisfying R3). If negative values are considered, the minimum is used instead of the maximum in \eqnref{cusum} and drifts are triggered when $g_t$ is smaller than $\lambda$. In summary, CUSUM respects R1 as $g$ is updated along new incoming data. However, the identity functions $f_i$ and $\phi(f_i)$ do not break time dependency, not satisfying R2 and leading to inconsistencies in $g$. Lastly, R4 is not satisfied since $f_i$ overfits (memorizing the current observation) and $g$ underfits (too restrictive bias) data since a single cumulative linear model may not be enough to represent more complex behaviors.

The Page-Hinkley Test (PHT), also proposed by~\citet{page:bio:54}, is a variation of CUSUM (using the same window configuration) in the sense it assesses data changes in terms of standard deviation measures rather than using averages. Thus, given the average estimation ${\mu_t = 1/(t - s)\sum_{j = s}^t x(j)}$, where interval ${[s, t]}$ representing the evolution of some phenomenon from the start ($s$) to the current window ($t$), PHT reports a drift whenever:
\begin{equation}
g_t = \lVert m_t - M_t \rVert > \lambda,
\label{eqn:page-hinkley}
\end{equation}
where ${m_t = \sum_{k = s}^t (x(k) - \mu_k)}$, ${M_t = \min(m_{[s, t]})}$.

Therefore, a drift occurs whenever the difference between the cumulative standard deviation is $\lambda$ units greater than the minimum standard deviation observed up to the current moment. Similarly to CUSUM, model $g$ is updated as new windows are processed and a reset occurs in case a drift is issued, such that both requirements R1 and R3 are satisfied. However, once $m_t$ is computed over a time-dependent sequence of observations, then R2 is not respected. In addition, despite PHT is slightly more complex than CUSUM, it is still prone of overfitting (failing R4).

The Adaptive Sliding Window (ADWIN), proposed by~\citet{bifet:sigkdd:09}, also comprises an extension of CUSUM, but applied over different window configurations. More precisely, given a timestamp between the start and current time of some phenomenon, the data stream $\fold{D}$ is divided into two adaptive windows $W_{[s,k]} = \{x(s), \cdots, x(k)\}$ and $W_{[k+1,t]} = \{x(k + 1), \cdots, x(t)\}$, for each ${k \in [s, t]}$. In this context, ADWIN reports a drift whenever:
\begin{equation}
g_k = \lVert \mu_{W_{[s,k]}} - \mu_{W_{[s+1,t]}} \rVert_1 > \lambda, ~~ \forall ~ k \in [s,t),
\label{adwin}
\end{equation}
in which $\mu_{W_{a,b}}$ is the average of $W_{a,b}$. As soon as a drift is issued, then $s = t$ in order to reset the past model and start a new phenomenon (respecting R3). However, the algorithm just compares averages between consecutive windows, taking no advantage from past data to update $g$, thus not satisfying R1. Further, as $f_i$ was inferred directly from data stream observations, R2 is not respected either. Lastly, despite the search space of $g$ is greater than the ones considered by CUSUM and PHT (since more windows are taken into account), the usage of an average model $f_i: W_{[a,b]} \rightarrow \mu_{W_[a,b]}$ and the fact that $g$ is too simplistic are still prone to underfitting.

From a different perspective, \citet{vallim:eswa:14} proposed the Unidimensional Fourier Transform (UDFT) to infer a model $f_i: C_i \rightarrow C_i$, based on the set of Fourier coefficients~\citep{bracewell:78:book} ${C_i = \{c_{in}, \cdots, c_{in + n-1}\}}$, as follows:
\begin{equation}
c_{in + j} = \dfrac{\epsilon}{n} \sum_{k = 0}^{n - 1} x_{in + k} e^{-\vecx{i} j 2\pi \frac{k}{n-1}}, \quad \epsilon = \left\{
\begin{array}{cc}
1, ~ & j = 0, \\
2, ~ & j > 0,
\end{array} \right.
\label{eqn:discrete-fourier-transform}
\end{equation}
where ${0 \leq j \leq (n-1)/2}$ and $\vecx{i}$ is the imaginary unit typically used to express the imaginary component of complex numbers, such that $e^{\vecx{i}x} = \cos(x) + \vecx{i}\sin(x)$. Moreover, we define $\vecx{i}$ in bold since the variable $i$ was representing the window index. In this context, $\phi(f_i)$ is the identify function and $g$ reports a drift whenever:
\begin{equation}
g_t = \lVert C_{t-1} - C_{t}\rVert_2 > \lambda,
\label{eqn:udef}    
\end{equation}
from which we conclude R1 is not satisfied since $g$ simply compares two consecutive windows, so that nothing is learned from past data. Furthermore, R3 is automatically respected since $g$ requires no reset. Moreover, the method is in accordance with R2, as Fourier coefficients are independent from each other. Lastly, this proposal is less prone to underfitting as the Fourier coefficients better represent data than averages and standard deviations. However, despite improving R4, such requirement is not fulfilled as $f_i$ still memorizes data and $g$ is ambiguous, since completely different coefficients may lead to similar Euclidean distances.

Later, \citet{costa:eswa:16} proposed the Cross Recurrence Concept Drift Detection (CRCDD) algorithm, which compares reconstructed phase spaces $\fold{P}_{t-1}, \fold{P}_{t}$ of consecutive windows in search for behavior changes using the Cross-Recurrence Analysis~\citep{marwan:pr:07,marwan:book:15}. Formally speaking, they analyze phase states using open balls with radius $\varepsilon$ in term of an $N \times N$ matrix $R$:
\begin{equation}
R_{a, b} = \left\{
\begin{array}{cl}
1, & ~ \text{if $\rho_a \in \fold{P}_{t-1}$ is a neighbor of $\rho_b \in \fold{P}_t$} \\
& ~ \text{according to an open ball centered at $\rho_a$} \\
& ~ \text{with radius $\varepsilon$}, \\
0, & ~ \text{otherwise},
\end{array} \right.
\label{crcdd}
\end{equation}
indicating when phase states $\rho_a \in \fold{P}_{t-1}$ and $\rho_b \in \fold{P}_t$ are close enough to each other (given the open ball). Later, they compute the Maximum Diagonal Length (MDL), \emph{\ie} the diagonal with maximum length represented by consecutive values equal to $1$ in $R$. In summary, $f_i: \fold{P}_i \rightarrow \fold{P}_i$ (respects R2), $\phi(f_i)$ is the identify function, and $g$ is in the form:
\begin{equation}
g_t = \text{MDL}(R) > \lambda.
\label{eqn:crcdd}    
\end{equation}

Therefore, R1 is not even considered as no knowledge is accumulated from past observations (R3 is automatically satisfied). R4 is partially satisfied for $g$ (since $f_i$ is the memory function), as MDL is computed using an open ball whose radius is set as the average of the maximum distances from all $\log(N)$-nearest neighbors to each phase state.

Comparisons among those methods in light of our methodology are detailed in \tabref{comparison}. Despite there are other algorithms in the literature~\citep[etc]{Gama2004,EDDM,6706768,Lev}, we could not include them due to the lack of space. 

\begin{table}[htb]
\renewcommand\arraystretch{1.2}
\setlength{\tabcolsep}{6pt}
\begin{center}
\caption{Comparison of concept drift methods regarding requirements R1--R4.}
\label{tab:comparison}
\begin{tabular}{|c|c|c|c|c|}
\hline
Method & Update (R1) & IID\ (R2) & Fixed JPD (R3) & BVD($f_i$, $g$) (R4) \\ \hline
CUSUM & Yes & No & Yes & (No, No) \\ \hline
PHT & Yes & No & Yes & (No, No) \\ \hline
ADWIN & No & No & Yes & (No, No) \\ \hline
UDFT  & No & Yes  & Yes  & (No, No) \\ \hline
CRCDD & No & Yes & Yes  & (Yes, No) \\ \hline
\end{tabular}
\end{center}
\end{table}
As observed, CRCDD have the strongest learning guarantees, meaning its drifts are most likely to be the result of actual changes in data behavior rather than by chance. From this, we suggest all authors to revise their approaches in light of our methodology.

\section{Conclusions}
\label{sec:conclusions}

This paper proposes a methodology to overcome the complexity involved in labeling data streams and the lack of theoretical learning guarantees in the CD scenario. In this context, a CD algorithm should be built according to the following steps: (i) window observations should be somehow reconstructed into another space in order to ensure data independence and allow identically sampling. Among the alternatives, we suggest to map them into phase spaces, using Dynamical System tools, to automatically define spaces $\fold{P}_i$ and $\fold{Y}_i$, as discussed along this paper. (ii) Given features from $f_i$ are extracted using function $\phi$, then the indicator function $g$ must compare past against current features and, in case no drift is issued, it should be updated to improve the representation of the current phenomenon. (iii) Conversely, if a drift is confirmed, then model $g$ should be reset to start analyzing a new phenomenon based on another JPD. Finally, (iv) the biases of both $g$ and $f_i$ should respect the BVD to avoid under/overfitting.

As elaborated in \secref{cd-algorithms-according-to-our-methodology}, where we show related work methods in lights of our methodology, we observed that, despite the provided requirements are typically ``known'', they are not fulfilled in different steps. Nevertheless, despite it is relatively easy to adapt algorithms in terms of our requirements R1, R2 and R3 ($g$ is updated, input data is i.i.d. and $g$ disregards accumulated data when a novelty occurs, respectively), the algorithm bias (R4) cannot be changed in most of the cases (at least without changing the nature of the algorithm itself in the process). For example, the CUSUM is too simplistic, and even forcing it to respect R1, R2 and R3, the usage of the window average to report drifts goes in discordance to the Bias-Variance Dilemma. Moreover, our contribution also explains the \textbf{reasons} those requirements exist. As far as we observe in practice, even when other researches consider R1--R4, they do not know \textbf{why} they are doing so. In that sense, we have bring that to light by associating those topics with the SLT framework.

According to our methodology, the CRCDD provides the strongest learning guarantees. In that sense, this does not mean others will not work and generate fair results. Actually, this implies that CRCDD have stronger probabilistic convergences to keep reporting fair results for incoming windows (unseen observations). We expect such analysis to be helpful to other researchers who intend to design new CD algorithms or evaluate the existent ones. Lastly, after employing our strategy to ensure learning bounds, other performance measures can be safely used to validate the quality of reported drifts, such as the of the MTBFA, MTD and MDR.

\section*{Acknowledgements}
We acknowledge sponsorships of FAPESP (S\~{a}o Paulo Research Foundation) and CNPq (National Counsel of Technological and Scientific Development).  Any opinions, findings, and conclusions or recommendations expressed in this material are those of the authors and do not necessarily reflect the views of FAPESP nor CNPq.

\section*{Disclosure statement}
The authors declare that they have no conflict of interest.

\section*{Funding}
This research was supported by FAPESP and CNPq, grants 2018/10652-9, 2017/16548-6 and 302077/2017-0.

\bibliographystyle{abbrvnat}
\setcitestyle{authoryear}
\bibliography{bibliography}

\begin{thebibliography}{43}
\providecommand{\natexlab}[1]{#1}
\providecommand{\url}[1]{\texttt{#1}}
\expandafter\ifx\csname urlstyle\endcsname\relax
  \providecommand{\doi}[1]{doi: #1}\else
  \providecommand{\doi}{doi: \begingroup \urlstyle{rm}\Url}\fi

\bibitem[Agarwal(1995)]{agarwal1995dynamical}
R.~Agarwal.
\newblock \emph{Dynamical Systems and Applications}, volume~4 of \emph{World
  Scientific series in applicable analysis}.
\newblock World Scientific, Singapore, Malaysia, 1995.

\bibitem[Alligood et~al.(1996)Alligood, Sauer, and Yorke]{alligood:96:book}
K.~T. Alligood, T.~Sauer, and J.~A. Yorke.
\newblock \emph{Chaos : an introduction to dynamical systems}.
\newblock Textbooks in mathematical sciences. Springer, 1996.

\bibitem[Andrievskii and Fradkov(2003)]{Andrievskii2003}
B.~R. Andrievskii and A.~L. Fradkov.
\newblock Control of chaos: Methods and applications. i. methods.
\newblock \emph{Automation and Remote Control}, 64\penalty0 (5):\penalty0
  673--713, 2003.

\bibitem[Bache and Lichman(2013)]{lichman:uci:13}
K.~Bache and M.~Lichman.
\newblock {UCI} machine learning repository, 2013.
\newblock URL \url{http://archive.ics.uci.edu/ml}.

\bibitem[Baena-Garc{\'{\i}}a et~al.(2006)Baena-Garc{\'{\i}}a, del
  Campo-{\'{A}}vila, Fidalgo, Bifet, Gavald{\'{a}}, and Morales-Bueno]{EDDM}
M.~Baena-Garc{\'{\i}}a, J.~del Campo-{\'{A}}vila, R.~Fidalgo, A.~Bifet,
  R.~Gavald{\'{a}}, and R.~Morales-Bueno.
\newblock {Early drift detection method}, 2006.

\bibitem[Bifet et~al.(2009)Bifet, Holmes, Pfahringer, Kirkby, and
  Gavald\`{a}]{bifet:sigkdd:09}
A.~Bifet, G.~Holmes, B.~Pfahringer, R.~Kirkby, and R.~Gavald\`{a}.
\newblock New ensemble methods for evolving data streams.
\newblock In \emph{Proc. 15th ACM SIGKDD International Conference on Knowledge
  Discovery and Data Mining}, KDD '09, pages 139--148, Paris, France, 2009.
  ACM.

\bibitem[Bifet et~al.(2010)Bifet, Holmes, and Pfahringer]{Lev}
A.~Bifet, G.~Holmes, and B.~Pfahringer.
\newblock Leveraging bagging for evolving data streams.
\newblock In \emph{Machine Learning and Knowledge Discovery in Databases,
  European Conference}, pages 135--150, Barcelona, Spain, 2010. Springer Berlin
  Heidelberg.

\bibitem[Bousquet and Elisseeff(2002)]{bousquet:jlmr:02}
O.~Bousquet and A.~Elisseeff.
\newblock Stability and generalization.
\newblock \emph{J. Mach. Learn. Res.}, 2:\penalty0 499--526, Mar. 2002.

\bibitem[Bracewell(1978)]{bracewell:78:book}
R.~Bracewell.
\newblock \emph{{The fourier transform and its applications}}.
\newblock McGraw-Hill Kogakusha, Ltd., second edition, 1978.

\bibitem[da~Costa et~al.(2016)da~Costa, Rios, and de~Mello]{costa:eswa:16}
F.~G. da~Costa, R.~A. Rios, and R.~F. de~Mello.
\newblock Using dynamical systems tools to detect concept drift in data
  streams.
\newblock \emph{Expert Systems with Applications}, 60:\penalty0 39 -- 50, 2016.

\bibitem[da~Costa et~al.(2017)da~Costa, Duarte, Vallim, and
  de~Mello]{costa:eswa:17}
F.~G. da~Costa, F.~S. L.~G. Duarte, R.~M.~M. Vallim, and R.~F. de~Mello.
\newblock Multidimensional surrogate stability to detect data stream concept
  drift.
\newblock \emph{Expert Syst. Appl.}, 87:\penalty0 15--29, 2017.

\bibitem[de~Mello and Moacir(2018)]{mello:book:18}
R.~F. de~Mello and A.~P. Moacir.
\newblock \emph{{A Practical Approach on the Statistical Learning Theory}},
  volume~1.
\newblock Springer, New York, USA, 2018.

\bibitem[de~Mello et~al.(2019)de~Mello, Vaz, Grossi, and Bifet]{mello:yuli:18}
R.~F. de~Mello, Y.~Vaz, C.~H. Grossi, and A.~Bifet.
\newblock On learning guarantees to unsupervised concept drift detection on
  data streams.
\newblock \emph{Expert Systems with Applications}, 117:\penalty0 90 -- 102,
  2019.

\bibitem[Devroye et~al.(1996)Devroye, Gy{\"{o}}rfi, and
  Lugosi]{devroye:96:book}
L.~Devroye, L.~Gy{\"{o}}rfi, and G.~Lugosi.
\newblock \emph{{A Probabilistic Theory of Pattern Recognition}}.
\newblock Springer, New York, 1996.

\bibitem[Gama et~al.(2004{\natexlab{a}})Gama, Medas, Castillo, and
  Rodrigues]{Gama2004}
J.~Gama, P.~Medas, G.~Castillo, and P.~Rodrigues.
\newblock Learning with drift detection.
\newblock In \emph{Advances in Artificial Intelligence -- 17th Brazilian
  Symposium on Artificial Intelligence}, pages 286--295, Berlin, Heidelberg,
  2004{\natexlab{a}}. Springer Berlin Heidelberg.

\bibitem[Gama et~al.(2004{\natexlab{b}})Gama, Medas, and Rocha]{gama:sac:04}
J.~Gama, P.~Medas, and R.~Rocha.
\newblock Forest trees for on-line data.
\newblock In \emph{Proc. 2004 ACM Symposium on Applied Computing}, SAC '04,
  pages 632--636, Nicosia, Cyprus, 2004{\natexlab{b}}. ACM.

\bibitem[Gama et~al.(2014)Gama, \v{Z}liobait\.{e}, Bifet, Pechenizkiy, and
  Bouchachia]{gama:14:acm}
J.~Gama, I.~\v{Z}liobait\.{e}, A.~Bifet, M.~Pechenizkiy, and A.~Bouchachia.
\newblock {A survey on concept drift adaptation}.
\newblock \emph{ACM Comput. Surv.}, 46\penalty0 (4):\penalty0 44:1--44:37,
  2014.

\bibitem[Geman et~al.(1992)Geman, Bienenstock, and Doursat]{geman:92:nc}
S.~Geman, E.~Bienenstock, and R.~Doursat.
\newblock {Neural networks and the bias/variance dilemma}.
\newblock \emph{Neural Computation}, 4\penalty0 (1):\penalty0 1--58, 1992.

\bibitem[Haykin(2009)]{haykin:09:book}
S.~Haykin.
\newblock \emph{Neural Networks and Learning Machines}.
\newblock Number v. 10 in Neural networks and learning machines. Prentice Hall,
  2009.

\bibitem[Helmbold and Long(1994)]{helmbold:ml:1994}
D.~P. Helmbold and P.~M. Long.
\newblock Tracking drifting concepts by minimizing disagreements.
\newblock \emph{Machine Learning}, 14\penalty0 (1):\penalty0 27--45, Jan 1994.

\bibitem[Kantz and Schreiber(2003)]{kantz:97:book}
H.~Kantz and T.~Schreiber.
\newblock \emph{Nonlinear Time Series Analysis}.
\newblock Cambridge University Press, Cambridge, 2 edition, 2003.

\bibitem[Kuh et~al.(1991)Kuh, Petsche, and Rivest]{kuh:nips:1991}
A.~Kuh, T.~Petsche, and R.~L. Rivest.
\newblock Learning time-varying concepts.
\newblock In R.~P. Lippmann, J.~E. Moody, and D.~S. Touretzky, editors,
  \emph{Advances in Neural Information Processing Systems 3}, pages 183--189.
  Morgan-Kaufmann, 1991.

\bibitem[{Lu} et~al.(2018){Lu}, {Liu}, {Dong}, {Gu}, {Gama}, and
  {Zhang}]{lu:kde:18}
J.~{Lu}, A.~{Liu}, F.~{Dong}, F.~{Gu}, J.~{Gama}, and G.~{Zhang}.
\newblock Learning under concept drift: A review.
\newblock \emph{IEEE Transactions on Knowledge and Data Engineering}, pages
  1--1, 2018.

\bibitem[Luxburg and Sch\"{o}lkopf(2011)]{luxburg:11:book}
U.~V. Luxburg and B.~Sch\"{o}lkopf.
\newblock {Statistical learning theory: models, concepts, and results}.
\newblock In \emph{Inductive Logic}, volume~10 of \emph{Handbook of the History
  of Logic}, pages 651--706, Amsterdam, Netherlands, 2011. Elsevier.

\bibitem[Marwan and Webber(2015)]{marwan:book:15}
N.~Marwan and C.~L. Webber.
\newblock \emph{Mathematical and Computational Foundations of Recurrence
  Quantifications}, pages 3--43.
\newblock Springer International Publishing, Cham, 2015.

\bibitem[Marwan et~al.(2007)Marwan, Romano, Thiel, and Kurths]{marwan:pr:07}
N.~Marwan, M.~C. Romano, M.~Thiel, and J.~Kurths.
\newblock Recurrence plots for the analysis of complex systems.
\newblock \emph{Physics Reports}, 438\penalty0 (5–6):\penalty0 237 -- 329,
  2007.

\bibitem[McDiarmid(1989)]{mcdiarmid:sic:1989}
C.~McDiarmid.
\newblock \emph{On the method of bounded differences}, page 148–188.
\newblock London Mathematical Society Lecture Note Series. Cambridge University
  Press, 1989.

\bibitem[Metzger(1997)]{Metzger1997}
M.~A. Metzger.
\newblock Applications of nonlinear dynamical systems theory in developmental
  psychology: Motor and cognitive development.
\newblock \emph{Nonlinear Dynamics, Psychology, and Life Sciences}, 1\penalty0
  (1):\penalty0 55--68, 1997.

\bibitem[Page(1954)]{page:bio:54}
E.~S. Page.
\newblock {Continuous Inspection Schemes}.
\newblock \emph{Biometrika}, 41:\penalty0 100--115, 1954.

\bibitem[Pagliosa and de~Mello(2017)]{pagliosa:eswa:17}
L.~d.~C. Pagliosa and R.~F. de~Mello.
\newblock Applying a kernel function on time-dependent data to provide
  supervised-learning guarantees.
\newblock \emph{Expert Systems with Applications}, 71:\penalty0 216 -- 229,
  2017.

\bibitem[Ravindra and Hagedorn(1998)]{ravindra:1998}
B.~Ravindra and P.~Hagedorn.
\newblock {Invariants of chaotic attractor in a nonlinearly damped system}.
\newblock \emph{Journal of applied mechanics}, 65:\penalty0 875--879, 1998.

\bibitem[R\'{e}v\'{e}sz(1967)]{revesz:67:book}
P.~R\'{e}v\'{e}sz.
\newblock {The laws of large numbers}.
\newblock In \emph{Probability and Mathematical Statistics: A Series of
  Monographs and Textbooks}, pages 2--. Academic Press, Cambridge,
  Massachusetts, 1967.

\bibitem[Rios et~al.(2015)Rios, Parrott, Lange, and de~Mello]{RIOS201511}
R.~A. Rios, L.~Parrott, H.~Lange, and R.~F. de~Mello.
\newblock Estimating determinism rates to detect patterns in geospatial
  datasets.
\newblock \emph{Remote Sensing of Environment}, 156:\penalty0 11 -- 20, 2015.

\bibitem[Serr{\`a} et~al.(2009)Serr{\`a}, Serra, and Andrzejak]{serra:09:njp}
J.~Serr{\`a}, X.~Serra, and R.~G. Andrzejak.
\newblock {Cross recurrence quantification for cover song identification}.
\newblock \emph{New Journal of Physics}, 11:\penalty0 093017, 2009.

\bibitem[Takens(1981)]{takens:1981}
F.~Takens.
\newblock {Detecting strange attractors in turbulence}.
\newblock In \emph{Dynamical systems and turbulence}, pages 366--381. Springer,
  Berlin, Heidelberg", 1981.

\bibitem[Tsymbal(2004)]{tsymbal:04}
A.~Tsymbal.
\newblock The problem of concept drift: Definitions and related work.
\newblock Technical report, 2004.

\bibitem[Tu(2010)]{tu:book:10}
L.~Tu.
\newblock \emph{An Introduction to Manifolds}.
\newblock Universitext. Springer New York, 2010.

\bibitem[Tucker(1999)]{tucker:1999}
W.~Tucker.
\newblock {The Lorenz attractor exists}.
\newblock \emph{Comptes Rendus de l'Académie des Sciences - Series I -
  Mathematics}, 328\penalty0 (12):\penalty0 1197 -- 1202, 1999.

\bibitem[Valiant(1984)]{valiant:acm:1984}
L.~G. Valiant.
\newblock A theory of the learnable.
\newblock \emph{Commun. ACM}, 27\penalty0 (11):\penalty0 1134--1142, Nov. 1984.

\bibitem[Vallim and De~Mello(2014)]{vallim:eswa:14}
R.~M.~M. Vallim and R.~F. De~Mello.
\newblock Proposal of a new stability concept to detect changes in unsupervised
  data streams.
\newblock \emph{Expert Syst. Appl.}, 41\penalty0 (16):\penalty0 7350--7360,
  2014.

\bibitem[Vapnik(1998)]{vapnik:98:book}
V.~N. Vapnik.
\newblock \emph{{Statistical learning theory}}.
\newblock Wiley, New York, 1 edition, 1998.

\bibitem[Wang et~al.(2013)Wang, Minku, Ghezzi, Caltabiano, Tino, and
  Yao]{6706768}
S.~Wang, L.~L. Minku, D.~Ghezzi, D.~Caltabiano, P.~Tino, and X.~Yao.
\newblock Concept drift detection for online class imbalance learning.
\newblock In \emph{The 2013 International Joint Conference on Neural Networks
  (IJCNN)}, pages 1--10, Dallas, TX, USA, 2013. IEEE.

\bibitem[Whitney(1936)]{whitney:1936}
H.~Whitney.
\newblock Differentiable manifolds.
\newblock \emph{Annals of Mathematics}, 37\penalty0 (3):\penalty0 645--680,
  1936.

\end{thebibliography}

\end{document}